\documentclass[runningheads]{llncs}
\usepackage{xcolor}
\usepackage{multirow}
\usepackage{graphicx}
\usepackage{soul}
\usepackage{float}
\usepackage{caption}
\usepackage{subcaption}
\usepackage[subtle,margins=normal,leading=normal]{savetrees}
\begin{document}
\title{Data Alchemy: Mitigating Cross-Site Model Variability Through Test Time Data Calibration}
\titlerunning{Data Alchemy}
\author{Abhijeet Parida$^{1,2}$, Antonia Alomar$^{3}$, Zhifan Jiang$^1$, \\Pooneh Roshanitabrizi$^1$, Austin Tapp$^1$, Mar\'{i}a J. Ledesma-Carbayo$^2$, Ziyue Xu$^4$,\\ Syed Muhammed Anwar$^{1,5}$,
Marius George Linguraru $^{1,5}$, Holger R. Roth$^4$}
\authorrunning{A. Parida et al.}
%
\institute{Children’s National Hospital, Washington, DC, USA \and Universidad Politécnica de Madrid, Madrid, Spain \and Universitat Pompeu Fabra, Barcelona, Spain \and Nvidia Corporation, Santa Clara, CA, USA \and  George Washington University, Washington, DC, USA}
\maketitle              
\begin{abstract}
Deploying deep learning-based imaging tools across various clinical sites poses significant challenges due to inherent domain shifts and regulatory hurdles associated with site-specific fine-tuning.
For histopathology, stain normalization techniques can mitigate discrepancies, but they often fall short of eliminating inter-site variations. Therefore, we present \textit{Data Alchemy}, an explainable stain normalization method combined with test time data calibration via a template learning framework to overcome barriers in cross-site analysis.  \textit{Data Alchemy}  handles shifts inherent to multi-site data and minimizes them without needing to change the weights of the normalization or classifier networks. Our approach extends to unseen sites in various clinical settings where data domain discrepancies are unknown. Extensive experiments highlight the efficacy of our framework in tumor classification in hematoxylin and eosin-stained patches. Our explainable normalization method boosts classification tasks' area under the precision-recall curve (AUPR) by 0.165, 0.545 to 0.710. Additionally, \textit{Data Alchemy} 
 further reduces the multisite classification domain gap, by improving the 0.710 AUPR  an additional 0.142, elevating classification performance further to 0.852, from 0.545. Our \textit{Data Alchemy} framework can popularize precision medicine with minimal operational overhead by allowing for the seamless integration of pre-trained deep learning-based clinical tools across multiple sites.
 
\end{abstract}
\section{Introduction}
In recent years, deep learning-based methods have performed well for various medical imaging analysis tasks such as disease diagnosis, classification, and segmentation \cite{anaya2021overview,sahiner2019deep}. 
However, according to the United States Food and Drugs Administration, there is no approval for artificial intelligence and machine learning-enabled medical devices in histopathology for the calendar year 2023 \cite{fda}. This suggests few of the developed methods are usable in a clinical setting -- particularly in histopathology, due to known challenges of generalizability and robustness across sites. Data and protocol variability across sites further hamper the approval of regulatory compliance for such tools \cite{cheng2021challenges,he2020deployment}. The typical approach to improve a model's performance and generalizability is to calibrate each model at every site before deployment \cite{laves2020well,rajaraman2022deep}. While effective in some circumstances, model weight calibration resulting in substantial parameter-related modifications necessitates regulatory re-approval. To overcome these challenges, we propose a different approach. Instead of performing weight calibration that would necessitate regulatory re-approval of the model, we perform data calibration/template learning 
 using \textit{Data Alchemy} to reduce the gap domain and hence, solve the generalizability problem at test time. To establish the efficacy of this approach, we address tumor classification in digital histopathology images. 

In histopathology, cells and tissue samples must be stained to be visible under a microscope. Then, they are digitized using microscopic scanners. The resulting samples' appearance varies depending on several factors such as the used reagents, staining procedure, and scanner specifications. Such variations directly affect analysis performed both by a pathologist or automated classification algorithms \cite{SALVI2021104129}. Stain normalization has been investigated as a pre-processing step to reduce color variations between histopathology samples. This involves transferring the color (stain) of a source histology patch to a target patch, while preserving the morphological tissue structure (content). Several studies have shown that data augmentation and stain normalization help increase the prediction accuracy \cite{cancers15051503,Tellez2019QuantifyingTE,Voon2023}. However, striking the appropriate balance between structure preservation and color consistency is challenging, as the resulting samples either contain artifacts and hallucinations or suffer in color appearance.  

\noindent \textbf{Related Works:} 
 Conventional stain normalization methods are mostly based on histogram transformations or color deconvolution (stain separation) \cite{Haub2015,HOQUE2024101997}. Histogram transform-based methods usually impose the color characteristics of a reference patch to another source patch using linear transformations \cite{Reinhard_2001,Zheng_2018}. Color deconvolution is a method for decoupling light-absorbance and stain concentration in each pixel using spectral characteristics of different stains \cite{Rabinovich_2003}. In other works, such as \cite{Macenko_2009}, RGB images are transformed into optical flows for estimating the stain vectors using singular value decomposition (SVD). However, these methodologies tend to generate artifacts in the background and/or color discontinuities in the normalized images.

Recent efforts have focused on deep learning-based methods, especially those using generative adversarial networks (GANs). GAN-based methods target stain normalization as a style-transfer problem \cite{style_transfer}. Some proposed methods have used cycle-consistent generative adversarial networks (cycleGAN) to match the target distribution \cite{shaban2019staingan,StainNet,runz2021normalization}.
In another approach, content was disentangled from style, opening the possibility of multiple stain representations and, in classification tasks, outperforming conventional color augmentation techniques \cite{HistAuGAN}. However, GANs are computationally expensive, are prone to mode collapse, and can lead to undesired changes in the underlying morphological structures \cite{HOQUE2024101997}. 

\noindent \textbf{Our Contribution:} 1) A \textbf{stain normalization} method that combines the advantages of Singular Value Decomposition (SVD) transformations in the latent space with the non-linearity of convolutional networks to ensure structure preservation in a simple, interpretable, and computationally efficient manner.
 2) We propose a \textbf{test time data calibration} method via template learning called \textit{Data Alchemy} that improves model generalizability without altering parameters during testing, thus maintaining regulatory compliance. 3) We demonstrate the effectiveness of our strategies by evaluating them on histopathological tumor classification data.

\section{Methods and Experimental Settings}
\subsection{ Explainable stain normalization}\label{sec:stain-norm}
\begin{figure}[t]
  \centerline{\includegraphics[width=1\linewidth]{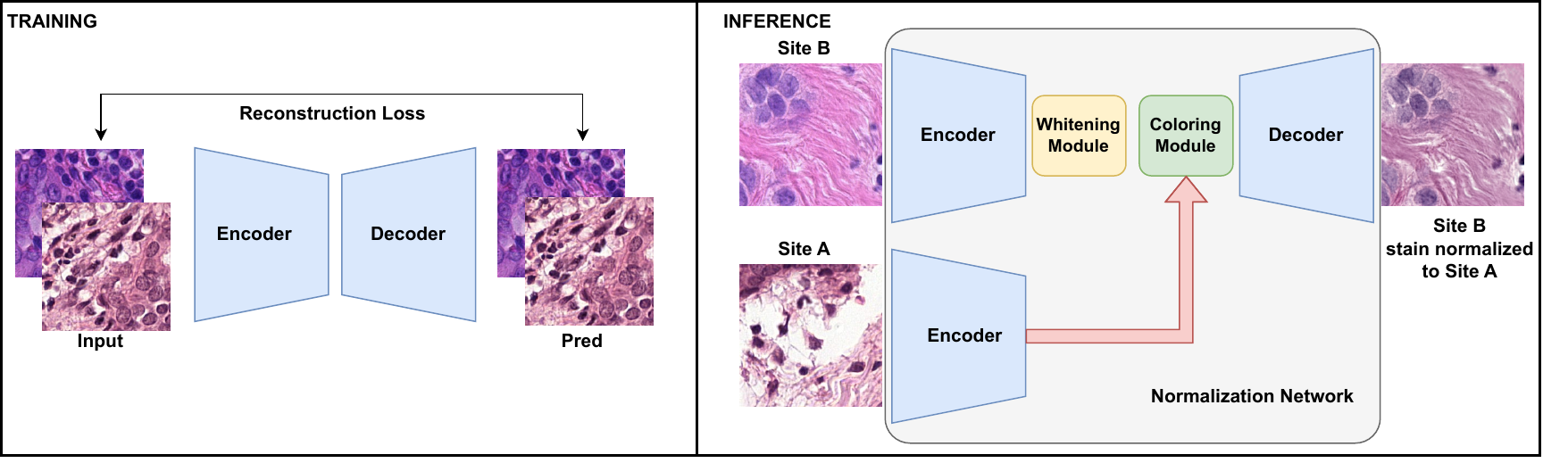}}
\caption{\textbf{Stain normalization} during training (left) and inference (right).}
\label{img:stain-normalization}
\end{figure}

We approached histopathology stain normalization as an image reconstruction task using feature transformation during inference, as shown in Fig.~\ref{img:stain-normalization}. 
Specifically, an image reconstruction network was trained using image $I$, such that $I\ = \ dec(enc(I))$,  where $enc(.)$ and $dec(.)$ are the encoder and decoder, respectively. Feature transformations were done using whitening and coloring transforms, proposed for arbitrary style transfer between natural images \cite{li2017universal}. 

The whitening transform was defined as $f_c\ =\ E_c\ D^{-\frac{1}{2}}_c\ E_c^T\ enc(I_c)$, where $D_c$ is a diagonal matrix of eigenvalues and $E_c$ is the orthogonal matrix of eigenvectors of the covariance matrix $enc(I_c) \cdot enc(I_c)^T$, and $I_c$ represent the patches that need to be re-stained. The covariance matrix is positive semi-definite, ensuring all eigenvalues $\geq 0$.  This whitening transform removed stain-specific information while preserving structure-related information from the patch that needs to be re-stained.

The ``whitened" $f_c$ was then ``colorized" using the coloring transforms, defined as $f_{cs}\ =\ E_s\ D^{\frac{1}{2}}_s\ E_s^T\ f_c$ \cite{hossain2014whitening,li2017universal}, where $D_s$ is a diagonal matrix of eigenvalues and $E_s$ is the orthogonal matrix of eigenvectors of the covariance matrix $enc(I_s) \cdot enc(I_s)^T$, and $I_s$ represents the patch whose staining parameters are used to stain the patch $I_c$. The coloring transform added stain-specific information from $I_s$ to the ``whitened" $f_c$.

The features of $I_c$ can be blended using a parameter $\alpha$ with re-stained features, $f_{cs}$, to control the stylization effect \cite{li2017universal}, as $f_{cs}\ =\ \alpha f_{cs} \ +\ (1-\alpha)\ enc(I_c)$. For patch staining, we set $\alpha=1$ as we aim to produce a stained patch and not control stylization.

\noindent \textbf{Implementation details:} We used all layers of VGG-19~\cite{vgg} upto `\textit{conv\_3\_3}' as the encoder and the exact inverted architecture of the encoder as the decoder (Fig.~\ref{img:stain-normalization}). We minimized L1 as a reconstruction loss using the AdamW optimizer for 10 epochs with a batch size of 96 and a learning rate of $1e^{-4}$. The best-performing model on the validation set was saved for stain normalization.

\subsection{Downstream classification task}\label{sec:clf}
To evaluate our stain normalization method, we reimplemented a downstream classification task from \cite{ncrf}. We used the ResNet-34 to identify tumor cells in small patches of whole slide images (WSIs). The classifier was trained using patches from one site, and tested on the unseen site. We compare the ResNet's accuracy using stain normalization with a fixed patch template and our proposed test time data calibration to establish generalizability.

\noindent \textbf{Implementation details:} We trained three models (ResNet-34): one on site, A, one on site B, and one on combined sites (A and B). The models were trained with augmentations from \cite{ncrf}, which included color jitter, changes in brightness, hue and saturation, random flips, and rotation. We minimized the cross-entropy loss for tumor \textit{vs.} healthy patches using the AdamW optimizer for 60 epochs with a batch size  256 and a learning rate of $1e^{-4}$. The model with the best validation metrics was chosen as our classifier.

\subsection{Data Alchemy: Test time data calibration}
\begin{figure}[t]
\centerline{
\includegraphics[width=0.7\linewidth]{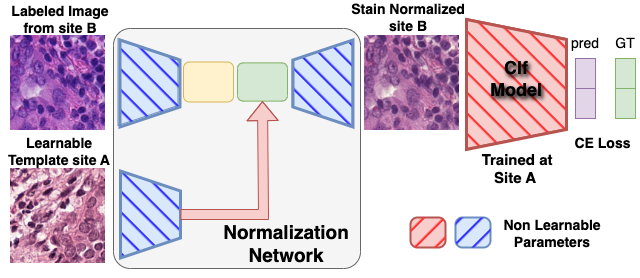}}
\caption{\textbf{Data Alchemy} uses the normalization network and a classifier to learn a template at test time to improve the classifier performance when deployed at a site.}
\label{img:calibration}
\end{figure}
For classifiers trained in Section \ref{sec:clf} to function optimally at different sites, a calibration step was necessary. We used the normalization method from Section \ref{sec:stain-norm} to adjust incoming patches to familiar stain parameters. Since we did not want to alter model weights, we propose adjusting the target template of the normalization network instead.

We randomly drew a real patch from the classifier's training site to instantiate a template. We froze the normalization network and the classifier and set the template tensor as learnable. During calibration, labeled images from the test site were normalized to match the staining of the training site. These stain-normalized images were passed through the classifier to obtain class logits, which were used to calculate the losses with the labels of the patches. The only learnable parameter was the template, so gradients were calculated for it, and over multiple iterations, the optimizer learned a synthetic template. Thus, the classifier guided the normalization network in modifying the template to improve the site's classification accuracy. This calibration step, was performed during deployment and is the test-time calibration called \textit{Data Alchemy}. The schematic for test-time calibration is shown in Fig.~\ref{img:calibration}.

\noindent \textbf{Implementation details:} The validation set from the sites is used to learn a template for calibrating the classifier. Half of the dataset was used to learn the template and the rest is the validation set of the data calibration step. Optimization is performed for 10 epochs to minimize the cross-entropy loss using the AdamW optimizer with a batch size of 256 and a learning rate of $1e^{-4}$.

\subsection{Dataset}

We used CAMELYON 16~\cite{bejnordi2017diagnostic}, a public dataset consisting of 400 WSIs of sentinel lymph nodes from two sites, site A - Radboud University Medical Center, Nijmegen, and site B -  University Medical Center,  Utrecht. Further, we used coordinates provided by 
Baidu Research~\cite{ncrf} to determine the presence or absence of tumor cells in 256x256 patches. The site-wise sample distribution is presented in Table~\ref{tab:dataset}.

\begin{table}[t]
\caption{\textbf{Data summary} of the splits, sites, and number of patches for the CAMELYON 16~\cite{bejnordi2017diagnostic}. The WSI for each of the sites is available in appendix \ref{app:data-splits}.}
\centering
\scriptsize
\resizebox{0.6\columnwidth}{!}{%
\begin{tabular}{l|cc|cc}
\cline{2-5}
           & \multicolumn{2}{c|}{\textbf{Site A}} & \multicolumn{2}{c}{\textbf{Site B}} \\ \cline{2-5} 
           & healthy           & tumor            & healthy           & tumor           \\ \hline
Train      & 77,204            & 52,000           & 71,447            & 60,000          \\
Validation & 33,306            & 58,000           & 11,551            & 8,000           \\
Test       & 20,183            & 34,000           & 8,449             & 12,000          \\ \hline
\end{tabular}
}\label{tab:dataset}
\end{table}

\subsection{Evaluation metrics} 
\textbf{Stain normalization:} We used metrics of structural similarity index measure (SSIM) and peak-signal-to-noise ratio (PSNR)~\cite{setiadi2021psnr} for evaluation. We also used specialized metrics \textit{cycleL1} and $AP(i, p)$~\cite{parida2024quantitative} to quantify the preservation of structural information and the accuracy of stain normalization. $cycleL1\ =\|I_c,\ sty(sty(I_c,\ I_s),\ I_c)\|$, is the norm between the original patch $I_c$ and the reconstructed original patch after two stain normalizations using normalization network $sty(.)$.  $Sty(.)$ stains $I_c$ to the parameters of $I_s$, from another site. This stained patch is re-stained with the staining parameters of $I_c$ to get the reconstructed original patch. For WSI, we adapt $AP(i, p)$ to measure changes in boundaries within patches highlighted using Sobel filters \cite{996}. For ideal stain normalization, \textit{cycleL1}  be 0 and $AP(i, p)$ should be 1.

\noindent \textbf{Tumor classification:} 
We used the area under the precision-recall (AUPR) curve and the area under the receiver operating characteristics (AUROC) curve as metrics.  Additionally, we reported a $F1$ score using the best threshold from the precision-recall curve.

\section{Results}
\subsection{Comparison with other stain normalization techniques}
\begin{figure}[t]
  \centerline{\includegraphics[width=1\linewidth]{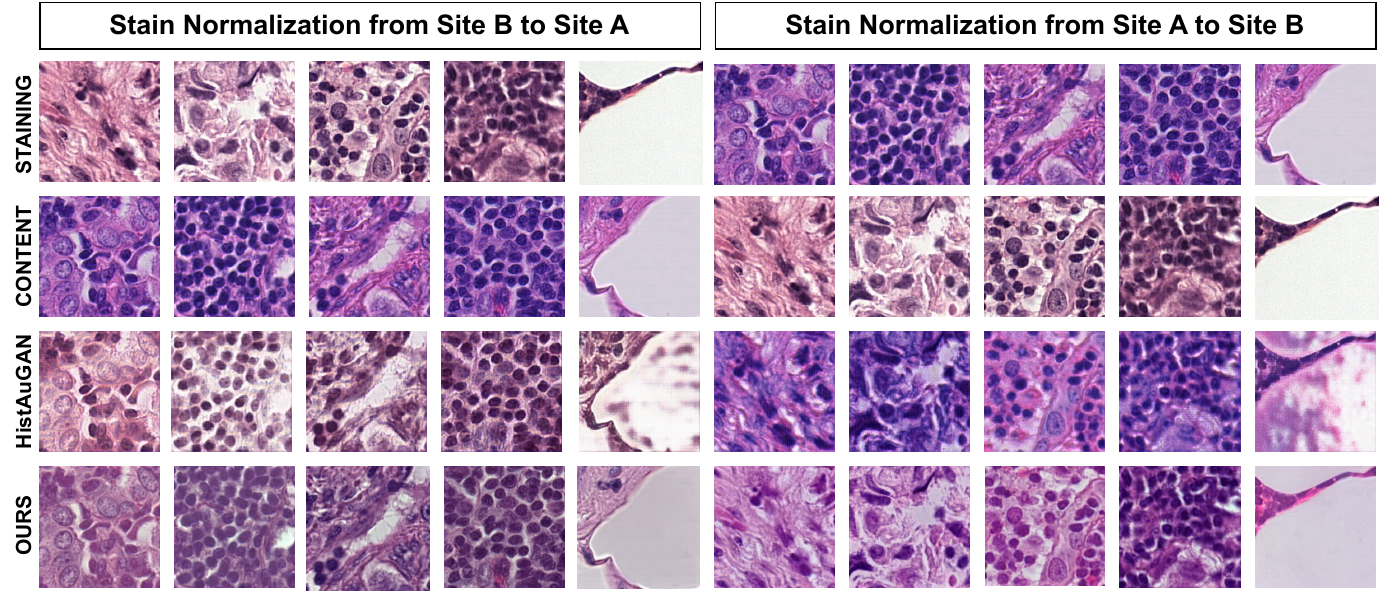}}
\caption{\textbf{Qualitative visualization} of stain normalization. 
A detailed version is available in appendix~\ref{app:stain_norm-comp}.}
\label{img:qualitative}
\end{figure}

For \textit{Data Alchemy}, the staining method must be controllable and capable of handling unseen stains during testing. So, we compare the performance of our proposed stain normalization method with HistAuGAN~\cite{HistAuGAN}, both quantitatively and qualitatively.  Fig.~\ref{img:qualitative} shows examples of stain normalization on different patches. Both approaches reduce the color appearance variations and create plausible stained samples while preserving the general structure visible in the original patches. However, HistAuGAN does not preserve the exact structures present in the original patch. It hallucinates additional nuclei and generates artifacts in the white background (Appendix~\ref {app:stain_norm-comp}). In contrast, our method preserves structural details better without any hallucinations or artifacts.

\begin{table}[b]
\caption{\textbf{Quantitave performance} comparison between various stain normalization methods. The reported numbers are average $\pm$ standard deviation.} \label{tab2:quantitative_performance}
\centering
\scriptsize
\resizebox{0.8\textwidth}{!}{%
\begin{tabular}{c|cccc}
\cline{2-5}
\multicolumn{1}{l|}{\multirow{2}{*}{\textbf{}}} &
  \multicolumn{1}{c|}{cycleL1} &
  \multicolumn{1}{c|}{SSIM} &
  \multicolumn{1}{c|}{PSNR} &
  AP(i, p) \\ \cline{2-5} 
\multicolumn{1}{l|}{} &
  \multicolumn{4}{c}{\textbf{A to B}} \\ \hline
HistAuGAN &
  \multicolumn{1}{c|}{0.060 $\pm$ 0.014} &
  \multicolumn{1}{c|}{0.691 $\pm$ 0.087} &
  \multicolumn{1}{c|}{14.893 $\pm$ 3.319} &
  \multicolumn{1}{c}{0.713 $\pm$ 1.123} \\
\textbf{Ours} &
  \multicolumn{1}{c|}{0.046 $\pm$ 0.018} &
  \multicolumn{1}{c|}{0.918 $\pm$ 0.060} &
  \multicolumn{1}{c|}{17.570 $\pm$ 3.061} &
  0.744 $\pm$ 0.269 \\ \hline
\textbf{} &
  \multicolumn{4}{c}{\textbf{B to A}} \\ \hline
HistAuGAN &
  \multicolumn{1}{c|}{0.062 $\pm$ 0.013} &
  \multicolumn{1}{c|}{0.720 $\pm$ 0.068} &
  \multicolumn{1}{c|}{14.465$\pm$ 2.714} &
  \multicolumn{1}{c}{0.553 $\pm$ 1.556} \\
\textbf{Ours} &
  \multicolumn{1}{c|}{0.043 $\pm$ 0.030} &
  \multicolumn{1}{c|}{0.896 $\pm$ 0.139} &
  \multicolumn{1}{c|}{18.827 $\pm$ 3.783} &
  0.845 $\pm$ 0.123 \\ \hline
\end{tabular}%
}
\end{table}

Table~\ref{tab2:quantitative_performance} shows that 
 our proposed stain normalization has a lower \textit{cycleL1} error compared to HistAuGAN. Moreover, our proposed method performs better in terms of \textit{SSIM} and \textit{PSNR}. 
These, together with higher values of $AP(i,p)$ and the qualitative examples, suggest that our proposed method is better at preserving the structural information present in the original patch, hence a better choice for stain normalization. 
Therefore, the subsequent classification tasks are performed using our proposed stain normalization module.  

\begin{figure}[t]
  \centerline{\includegraphics[width=0.9\linewidth]{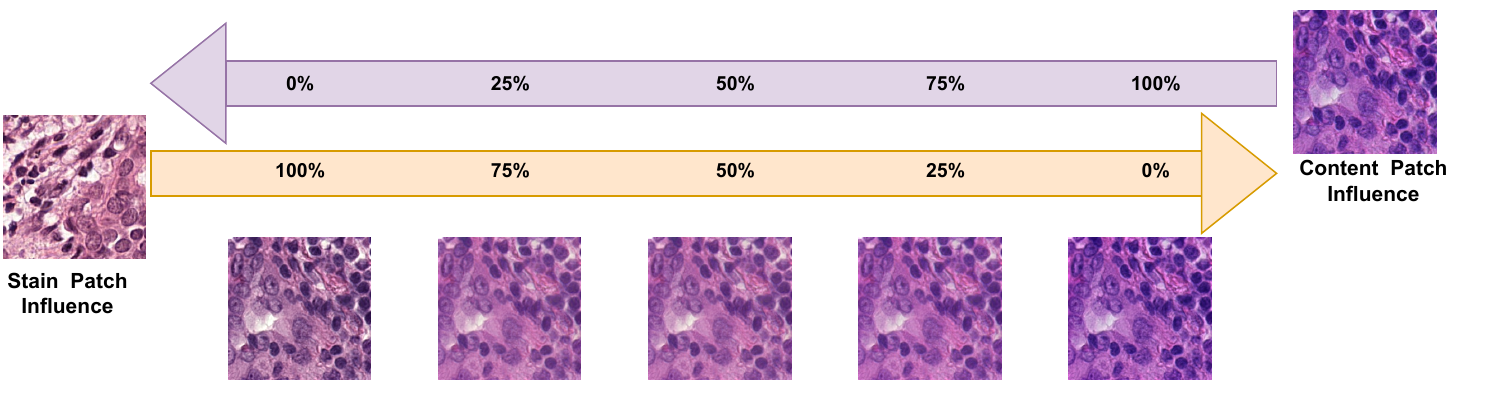}}
\caption{\textbf{Eigenvalue blending} of the content and stain patches produces different staining. The (\%) in the arrows are the content and stain patches blending weights.}
\label{img:eigen-blend}
\end{figure}

\noindent{\textbf{Exploring explainability in stain normalization:}}
\begin{figure}[t]
  \centerline{\includegraphics[width=0.7\linewidth]{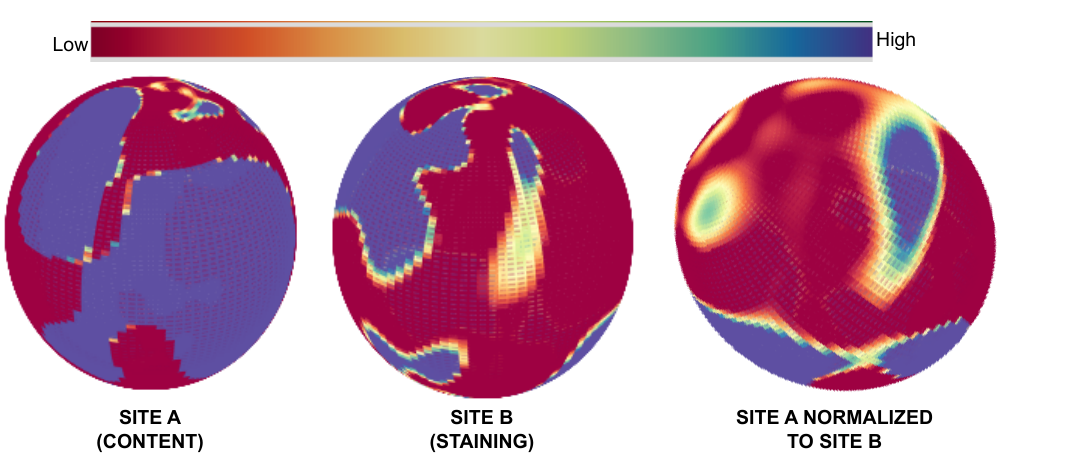}}
\caption{\textbf{Eigenvalue and vectors } of a content, stain, and stain normalized patch projected onto a 3D sphere. Each point of the surface represents an eigenvector and the color represents an eigenvalue. Red signifies a smaller and blue higher eigenvalues. }
\label{img:eignevalues}
\end{figure}
In Fig.~\ref{img:eignevalues},
we show an example of the normalization of site A to site B. We can see that post normalization the higher eigenvalues of site A become smaller with lower eigenvalues of site B. Also, site A normalized to site B looks much closer to site B than to site A. 
 We hypothesize that eigenvalues and vector manipulation are sufficient in stain blending, as shown in Fig.~\ref{img:eigen-blend}. We can control the blending of two patches directly by purely using eigenvalues and vectors from two different sites. By controlling the effect of the eigenvalues and vectors, we can control the staining to one particular site or the other. The manipulation of the eigenvalues helps us understand why one site is stained in a particular way compared to the other. 

\subsection{Stain normalization on downstream task} 
\begin{table}[t]
\caption{\textbf{Quantitative performance} of tumor and non-tumor classification of patches in various settings. UBM and LBM are expected Upper Bound Model and Lower Bound Model performance when trained with one dataset only. Train, Norm, and Temp refer to the train data site, normalization, and normalization template site, respectively.}
\centering
\scriptsize

\resizebox{\columnwidth}{!}{%
\begin{tabular}{lccccccccccc}
\cline{1-12}
\multicolumn{1}{l|}{\multirow{2}{*}{}} &
  \multicolumn{1}{c|}{\multirow{2}{*}{Norm}} &
  \multicolumn{5}{c|}{\textbf{Testing on Site B}} &
  \multicolumn{5}{c}{\textbf{Testing on Site A}} \\ \cline{3-12} 
\multicolumn{1}{l|}{}           & \multicolumn{1}{c|}{}    & Train & Temp   & AUPR  & AUROC & \multicolumn{1}{c|}{F1}    & Train & Temp   & AUPR  & AUROC & F1    \\ \hline
\multicolumn{1}{l|}{Best Case}  & \multicolumn{1}{c|}{No}  & A, B  & -      & 0.939 & 0.904 & \multicolumn{1}{c|}{0.850} & A, B  & -      & 0.939 & 0.912 & 0.878 \\
\multicolumn{1}{l|}{UBM}        & \multicolumn{1}{c|}{No}  & B     & -      & 0.916 & 0.888 & \multicolumn{1}{c|}{0.840} & A     & -      & 0.930 & 0.881 & 0.848 \\
\multicolumn{1}{l|}{LBM}        & \multicolumn{1}{c|}{No}  & A     & -      & 0.545 & 0.525 & \multicolumn{1}{c|}{0.771} & B     & -      & 0.841 & 0.774 & 0.787 \\ \hline
                                &                          & \multicolumn{10}{c}{\textbf{Stain Normalization}}                                                    \\ \hline
\multicolumn{1}{l|}{1 Template} & \multicolumn{1}{c|}{Yes} & A     & A      & 0.710 & 0.704 & \multicolumn{1}{c|}{0.792} & B     & B      & 0.648 & 0.578 & 0.771 \\
\multicolumn{1}{l|}{\begin{tabular}[c]{@{}l@{}}10  Template\\ (Ensemble)\end{tabular}} &
  \multicolumn{1}{c|}{Yes} &
  A &
  A &
  0.672 &
  0.689 &
  \multicolumn{1}{c|}{0.799} &
  B &
  B &
  0.686 &
  0.630 &
  0.773 \\
\multicolumn{1}{l|}{\begin{tabular}[c]{@{}l@{}}1 Template\\ (Test Data)\end{tabular}} &
  \multicolumn{1}{c|}{Yes} &
  A &
  B &
  0.587 &
  0.591 &
  \multicolumn{1}{c|}{0.782} &
  B &
  A &
  0.726 &
  0.652 &
  0.780 \\ \hline
                                &                          & \multicolumn{10}{c}{\textbf{Data Alchemy: Test Time Data Calibration}}                               \\ \hline
\multicolumn{1}{l|}{}           & \multicolumn{1}{c|}{Yes} & A     & Learnt & 0.852 & 0.833 & \multicolumn{1}{c|}{0.817} & B     & Learnt & 0.928 & 0.890 & 0.856 \\ \hline
\end{tabular}%
}
\label{tab:clf-1}
\end{table}
Table~\ref{tab:clf-1} shows that the classifier performs best when trained on data from both sites A and B. The upper bound model (UBM) represents the classifier trained and tested on the same data location, while the lower bound model (LBM) shows the performance drop (0.394 \textit{AUPR}) when testing on a site different from the training site. Stain normalization to a single template from site A improves classifier performance beyond the LBM when the classifier is trained on site A and tested on site B. Using one or ten templates increases the \textit{AUPR} scores by 0.165 and 0.127, respectively, over the LBM. When using a template from site B with a model trained on site A, we observe a negligible improvement of 0.042. This demonstrates that stain normalization improves classifier performance.

Table~\ref{tab:clf-1} also shows that in some scenarios, the LBM is close to the UBM, indicating that training on site B captures the necessary data diversity for good performance on site A. More visualizations of the phenomenon are in Appendix \ref{app:stain_norm}. So  When staining patches from site A to B, there is a drop of 0.193 and 0.155 AUPR using one or ten templates, respectively. A single template from site A only drops performance by 0.115. Overall, these findings suggest that static stain normalization may not always be beneficial for classifier performance.
  
\subsection{Test time data calibration}

Since static stain normalization may not guarantee optimal performance and we cannot update the model parameters due to regulatory concerns, we apply \textit{Data Alchemy} to the classifier. In Table~\ref{tab:clf-1}, the classifier trained on site A and tested on site B, the learned template boosts performance by 0.307 \textit{AUPR} over the LBM and is just 0.064 below the UBM. Additionally, the classifier trained on site B and tested on site A also improves performance by 0.087 \textit{AUPR} over the LBM and is only 0.002 below the UBM. We also observe an improvement of 0.009 AUPR and 0.008 F1 score of the data-calibrated model over the UBM. This demonstrates that \textit{Data Alchemy}'s learned template enhances classifier performance across different sites and has the potential to surpass the UBM.
 
\section{Conclusion}
We propose an effective and explainable stain normalization strategy that preserves the image structures and reduces stain variance between a template image and the original patch. Moreover, data calibration using \textit{Data Alchemy} improves the classification accuracy without retraining of any kind.  It serves as a step that enhances classifier generalizability, reducing the domain gap between multiple sites. Apart from easing regulatory approval hurdles, \textit{Data Alchemy} may be used for onsite model weight calibration when it is difficult to access the model (e.g., API-based interaction) or update the model (e.g., black boxes that do not support retraining or continuous learning).
\\
\\
\noindent \textbf{Acknowledgements}
\\
This work was supported by the National Cancer Institute award UG3CA236536.

\bibliographystyle{splncs04}
\bibliography{refs}

@Article{cancers15051503,
AUTHOR = {Salvi, Massimo and Caputo, Alessandro and Balmativola, Davide and Scotto, Manuela and Pennisi, Orazio and Michielli, Nicola and Mogetta, Alessandro and Molinari, Filippo and Fraggetta, Filippo},
TITLE = {Impact of Stain Normalization on Pathologist Assessment of Prostate Cancer: A Comparative Study},
JOURNAL = {Cancers},
VOLUME = {15},
YEAR = {2023},
NUMBER = {5},
ARTICLE-NUMBER = {1503},
URL = {https://www.mdpi.com/2072-6694/15/5/1503},
PubMedID = {36900293},
ISSN = {2072-6694},
DOI = {10.3390/cancers15051503}
}

@article{Voon2023,
   author = {Wingates Voon and Yan Chai Hum and Yee Kai Tee and Wun-She Yap and Humaira Nisar and Hamam Mokayed and Neha Gupta and Khin Wee Lai},
   doi = {10.1038/s41598-023-46619-6},
   issn = {2045-2322},
   issue = {1},
   journal = {Scientific Reports},
   month = {11},
   pages = {20518},
   title = {Evaluating the effectiveness of stain normalization techniques in automated grading of invasive ductal carcinoma histopathological images},
   volume = {13},
   year = {2023},
}

@article{vgg,
  title={Very deep convolutional networks for large-scale image recognition},
  author={Simonyan, Karen and Zisserman, Andrew},
  journal={arXiv preprint arXiv:1409.1556},
  year={2014}
}

@article{runz2021normalization,
author = {Runz, Marlen and Rusche, Daniel and Schmidt, Stefan and Weihrauch, Martin and Hesser, Jürgen and Weis, Cleo-Aron},
title = {Normalization of HE-stained histological images using cycle consistent generative adversarial networks},
journal = {Diagnostic Pathology},
year = {2021},
volume={16},
pages={71},
doi = {10.1186/s13000-021-01126-y}
}

@ARTICLE{StainNet,
  
AUTHOR={Kang, Hongtao and Luo, Die and Feng, Weihua and Zeng, Shaoqun and Quan, Tingwei and Hu, Junbo and Liu, Xiuli},   
	 
TITLE={StainNet: A Fast and Robust Stain Normalization Network},      
	
JOURNAL={Frontiers in Medicine},      
	
VOLUME={8},           
	
YEAR={2021},      
	  
URL={https://www.frontiersin.org/articles/10.3389/fmed.2021.746307},       
	
DOI={10.3389/fmed.2021.746307},      
	
ISSN={2296-858X}
}

@inproceedings{HistAuGAN,
  title={Structure-preserving multi-domain stain color augmentation using style-transfer with disentangled representations},
  author={Wagner, Sophia J and Khalili, Nadieh and Sharma, Raghav and Boxberg, Melanie and Marr, Carsten and de Back, Walter and Peng, Tingying},
  booktitle={Medical Image Computing and Computer Assisted Intervention--MICCAI 2021: 24th International Conference, Strasbourg, France, September 27--October 1, 2021, Proceedings, Part VIII 24},
  pages={257--266},
  year={2021},
  organization={Springer}
}

@article{Tellez2019QuantifyingTE,
  title={Quantifying the effects of data augmentation and stain color normalization in convolutional neural networks for computational pathology},
  author={David Tellez and Geert J. S. Litjens and P{\'e}ter B{\'a}ndi and Wouter Bulten and John-Melle Bokhorst and Francesco Ciompi and Jeroen van der Laak},
  journal={Medical image analysis},
  year={2019},
  volume={58},
  pages={
          101544
        },
  url={https://api.semanticscholar.org/CorpusID:62841444}
}

@inproceedings{ncrf,
    title={Cancer Metastasis Detection With Neural Conditional Random Field},
    booktitle={Medical Imaging with Deep Learning},
    author={Li, Yi and Ping, Wei},
    year={2018}
}

@article{sahiner2019deep,
  title={Deep learning in medical imaging and radiation therapy},
  author={Sahiner, Berkman and Pezeshk, Aria and Hadjiiski, Lubomir M and Wang, Xiaosong and Drukker, Karen and Cha, Kenny H and Summers, Ronald M and Giger, Maryellen L},
  journal={Medical physics},
  volume={46},
  number={1},
  pages={e1--e36},
  year={2019},
  publisher={Wiley Online Library}
}

@article{anaya2021overview,
  title={An overview of deep learning in medical imaging},
  author={Anaya-Isaza, Andr{\'e}s and Mera-Jim{\'e}nez, Leonel and Zequera-Diaz, Martha},
  journal={Informatics in medicine unlocked},
  volume={26},
  pages={100723},
  year={2021},
  publisher={Elsevier}
}

@inproceedings{laves2020well,
  title={Well-calibrated regression uncertainty in medical imaging with deep learning},
  author={Laves, Max-Heinrich and Ihler, Sontje and Fast, Jacob F and Kahrs, L{\"u}der A and Ortmaier, Tobias},
  booktitle={Medical Imaging with Deep Learning},
  pages={393--412},
  year={2020},
  organization={PMLR}
}

@article{rajaraman2022deep,
  title={Deep learning model calibration for improving performance in class-imbalanced medical image classification tasks},
  author={Rajaraman, Sivaramakrishnan and Ganesan, Prasanth and Antani, Sameer},
  journal={PloS one},
  volume={17},
  number={1},
  pages={e0262838},
  year={2022},
  publisher={Public Library of Science San Francisco, CA USA}
}

@article{cheng2021challenges,
  title={Challenges in the development, deployment, and regulation of artificial intelligence in anatomic pathology},
  author={Cheng, Jerome Y and Abel, Jacob T and Balis, Ulysses GJ and McClintock, David S and Pantanowitz, Liron},
  journal={The American Journal of Pathology},
  volume={191},
  number={10},
  pages={1684--1692},
  year={2021},
  publisher={Elsevier}
}

@article{he2020deployment,
  title={Deployment of artificial intelligence in real-world practice: opportunity and challenge},
  author={He, Mingguang and Li, Zhixi and Liu, Chi and Shi, Danli and Tan, Zachary},
  journal={The Asia-Pacific Journal of Ophthalmology},
  volume={9},
  number={4},
  pages={299--307},
  year={2020},
  publisher={LWW}
}

@article{bejnordi2017diagnostic,
  title={Diagnostic assessment of deep learning algorithms for detection of lymph node metastases in women with breast cancer},
  author={Bejnordi, Babak Ehteshami and Veta, Mitko and Van Diest, Paul Johannes and Van Ginneken, Bram and Karssemeijer, Nico and Litjens, Geert and Van Der Laak, Jeroen AWM and Hermsen, Meyke and Manson, Quirine F and Balkenhol, Maschenka and others},
  journal={Jama},
  volume={318},
  number={22},
  pages={2199--2210},
  year={2017},
  publisher={American Medical Association}
}

@INPROCEEDINGS{Macenko_2009,
  author={Macenko, Marc and Niethammer, Marc and Marron, J. S. and Borland, David and Woosley, John T. and Xiaojun Guan and Schmitt, Charles and Thomas, Nancy E.},
  booktitle={2009 IEEE International Symposium on Biomedical Imaging: From Nano to Macro}, 
  title={A method for normalizing histology slides for quantitative analysis}, 
  year={2009},
  volume={},
  number={},
  pages={1107-1110},
  doi={10.1109/ISBI.2009.5193250}}

@ARTICLE{Reinhard_2001,
  author={Reinhard, E. and Adhikhmin, M. and Gooch, B. and Shirley, P.},
  journal={IEEE Computer Graphics and Applications}, 
  title={Color transfer between images}, 
  year={2001},
  volume={21},
  number={5},
  pages={34-41},
  doi={10.1109/38.946629}}

@inproceedings{Rabinovich_2003,
 author = {Rabinovich, Andrew and Agarwal, Sameer and Laris, Casey and Price, Jeffrey and Belongie, Serge},
 booktitle = {Advances in Neural Information Processing Systems},
 editor = {S. Thrun and L. Saul and B. Sch\"{o}lkopf},
 pages = {},
 publisher = {MIT Press},
 title = {Unsupervised Color Decomposition Of Histologically Stained Tissue Samples},
 volume = {16},
 year = {2003}
}

@article{Haub2015,
   author = {Peter Haub and Tobias Meckel},
   doi = {10.1038/srep12096},
   issn = {2045-2322},
   issue = {1},
   journal = {Scientific Reports},
   month = {7},
   pages = {12096},
   title = {A Model based Survey of Colour Deconvolution in Diagnostic Brightfield Microscopy: Error Estimation and Spectral Consideration},
   volume = {5},
   year = {2015},
}

@ARTICLE{Zheng_2018,
  author={Zheng, Yushan and Jiang, Zhiguo and Zhang, Haopeng and Xie, Fengying and Ma, Yibing and Shi, Huaqiang and Zhao, Yu},
  journal={IEEE Transactions on Medical Imaging}, 
  title={Histopathological Whole Slide Image Analysis Using Context-Based CBIR}, 
  year={2018},
  volume={37},
  number={7},
  pages={1641-1652},
  doi={10.1109/TMI.2018.2796130}}

@article{HOQUE2024101997,
title = {Stain normalization methods for histopathology image analysis: A comprehensive review and experimental comparison},
journal = {Information Fusion},
volume = {102},
pages = {101997},
year = {2024},
issn = {1566-2535},
doi = {https://doi.org/10.1016/j.inffus.2023.101997},
url = {https://www.sciencedirect.com/science/article/pii/S1566253523003135},
author = {Md. Ziaul Hoque and Anja Keskinarkaus and Pia Nyberg and Tapio Seppänen}
}

@INPROCEEDINGS{style_transfer,
  author={Gatys, Leon A. and Ecker, Alexander S. and Bethge, Matthias},
  booktitle={2016 IEEE Conference on Computer Vision and Pattern Recognition (CVPR)}, 
  title={Image Style Transfer Using Convolutional Neural Networks}, 
  year={2016},
  volume={},
  number={},
  pages={2414-2423},
  doi={10.1109/CVPR.2016.265}}

@article{li2017universal,
  title={Universal style transfer via feature transforms},
  author={Li, Yijun and Fang, Chen and Yang, Jimei and Wang, Zhaowen and Lu, Xin and Yang, Ming-Hsuan},
  journal={Advances in neural information processing systems},
  volume={30},
  year={2017}
}

@article{hossain2014whitening,
  title={Whitening and coloring transformations for multivariate Gaussian data},
  author={Hossain, Maliha},
  journal={A slecture partly based on the ECE662 Spring},
  year={2014}
}

@article{parida2024quantitative,
  title={Quantitative Metrics for Benchmarking Medical Image Harmonization},
  author={Parida, Abhijeet and Jiang, Zhifan and Packer, Roger J and Avery, Robert A and Anwar, Syed M and Linguraru, Marius G},
  journal={arXiv preprint arXiv:2402.04426},
  year={2024}
}

@ARTICLE{996,
  author={Kanopoulos, N. and Vasanthavada, N. and Baker, R.L.},
  journal={IEEE Journal of Solid-State Circuits}, 
  title={Design of an image edge detection filter using the Sobel operator}, 
  year={1988},
  volume={23},
  number={2},
  pages={358-367},
  doi={10.1109/4.996}}

@article{setiadi2021psnr,
  title={PSNR vs SSIM: imperceptibility quality assessment for image steganography},
  author={Setiadi, De Rosal Igantius Moses},
  journal={Multimedia Tools and Applications},
  volume={80},
  number={6},
  pages={8423--8444},
  year={2021},
  publisher={Springer}
}

@article{SALVI2021104129,
title = {The impact of pre- and post-image processing techniques on deep learning frameworks: A comprehensive review for digital pathology image analysis},
journal = {Computers in Biology and Medicine},
volume = {128},
pages = {104129},
year = {2021},
issn = {0010-4825},
doi = {https://doi.org/10.1016/j.compbiomed.2020.104129},
url = {https://www.sciencedirect.com/science/article/pii/S0010482520304601},
author = {Massimo Salvi and U. Rajendra Acharya and Filippo Molinari and Kristen M. Meiburger}
}

@misc{fda, title={Artificial Intelligence and machine learning (AI/ml)-enabled medical Device}, url={https://www.fda.gov/medical-devices/software-medical-device-samd/artificial-intelligence-and-machine-learning-aiml-enabled-medical-devices}, journal={U.S. Food and Drug Administration}, publisher={FDA}, author={Center for Devices and Radiological Health, US FDA}, year={2023}, month={Dec}}

@inproceedings{shaban2019staingan,
  title={Staingan: Stain style transfer for digital histological images},
  author={Shaban, M Tarek and Baur, Christoph and Navab, Nassir and Albarqouni, Shadi},
  booktitle={2019 Ieee 16th international symposium on biomedical imaging (Isbi 2019)},
  pages={953--956},
  year={2019},
  organization={IEEE}
}
\newpage
\appendix
\section{Data splits}\label{app:data-splits}
\subsection{site A split json}
\{
    "site": "A", \\
    "val": [
        "tumor\_011.tif",
        "tumor\_047.tif",
        "tumor\_012.tif",
        "tumor\_028.tif",
        "tumor\_041.tif",
        "tumor\_045.tif",
        "tumor\_051.tif",
        "tumor\_053.tif",
        "tumor\_044.tif",
        "tumor\_016.tif",
        "tumor\_013.tif",
        "tumor\_042.tif",
        "tumor\_050.tif",
        "tumor\_021.tif",
        "tumor\_037.tif",
        "tumor\_014.tif",
        "tumor\_038.tif",
        "tumor\_043.tif",
        "tumor\_024.tif",
        "tumor\_036.tif",
        "tumor\_022.tif",
        "tumor\_019.tif",
        "tumor\_049.tif",
        "tumor\_039.tif",
        "tumor\_046.tif",
        "tumor\_032.tif",
        "tumor\_052.tif",
        "tumor\_040.tif",
        "tumor\_048.tif"
    ], \\
    "test": [
        "tumor\_068.tif",
        "tumor\_055.tif",
        "tumor\_058.tif",
        "tumor\_054.tif",
        "tumor\_057.tif",
        "tumor\_069.tif",
        "tumor\_063.tif",
        "tumor\_062.tif",
        "tumor\_056.tif",
        "tumor\_065.tif",
        "tumor\_061.tif",
        "tumor\_066.tif",
        "tumor\_070.tif",
        "tumor\_060.tif",
        "tumor\_064.tif",
        "tumor\_067.tif",
        "tumor\_059.tif"
    ]
\}

\subsection{site B split json}
\{
    "site": "B",\\
    "val": [
        "tumor\_104.tif",
        "normal\_142.tif",
        "normal\_148.tif",
        "tumor\_103.tif",
        "normal\_147.tif",
        "normal\_143.tif",
        "tumor\_102.tif",
        "normal\_141.tif",
        "normal\_150.tif",
        "tumor\_101.tif",
        "normal\_145.tif",
        "normal\_146.tif",
        "normal\_149.tif"
    ],\\
    "test": [
        "tumor\_108.tif",
        "normal\_157.tif",
        "normal\_151.tif",
        "normal\_155.tif",
        "normal\_156.tif",
        "tumor\_106.tif",
        "tumor\_109.tif",
        "tumor\_107.tif",
        "tumor\_110.tif",
        "normal\_158.tif",
        "normal\_153.tif",
        "normal\_159.tif",
        "normal\_154.tif",
        "tumor\_105.tif",
        "normal\_160.tif",
        "normal\_152.tif"
    ]
\}
\section{Visualize the stain normalization}\label{app:stain_norm}
 
\begin{figure}[H]
  \centerline{\includegraphics[width=0.8\linewidth]{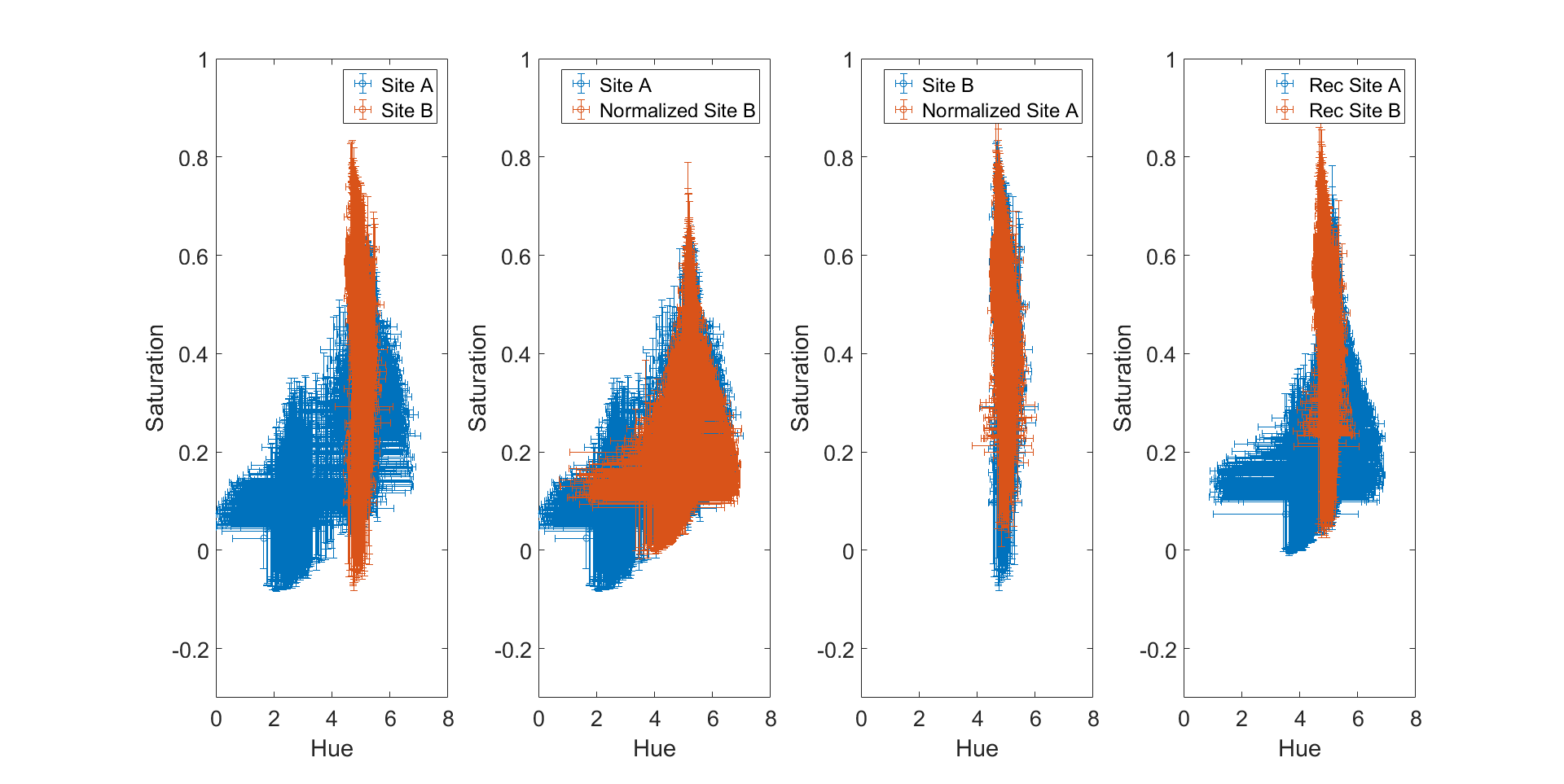}}
\caption{\textbf{HSV color space distribution.} Hue and Saturation mean and standard deviation distribution of the originals, stain normalized, and reconstructed patches. Site B might be a subset of site A in terms of Hue(H) and Saturation(V) space.}
\label{img:calibration-2}
\end{figure}
\newpage
\section{Comparative  stain normalization}\label{app:stain_norm-comp}
\begin{figure}[H]
  \centerline{\includegraphics[width=1.6\textwidth, angle=90]{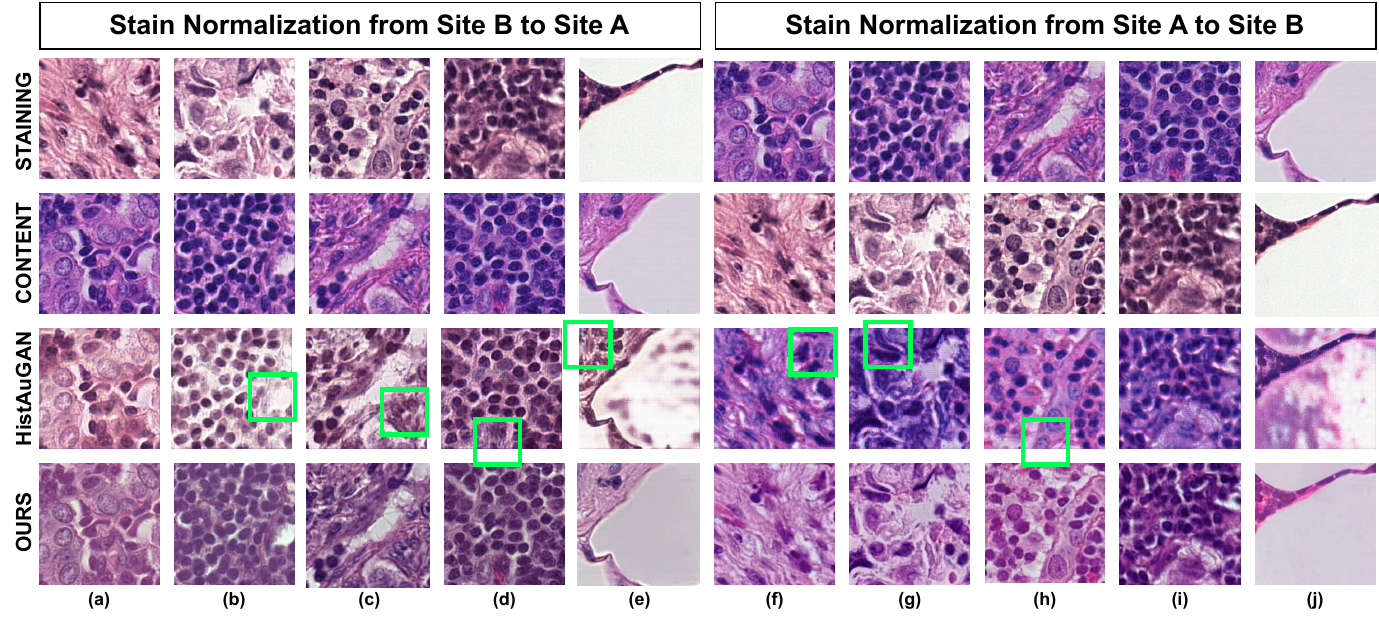}}
\caption{\textbf{Zoomed in qualitative prediction} from different stain normalizers. Green bounding boxes highlight GAN hallucination. We can appreciate in the examples e and j that when the patches contain white regions some artifacts appear in the stained regions of the HistAuGAN. }
\label{fig:app-zoom}
\end{figure}

\end{document}